\pdfoutput=1

\documentclass[11pt]{article}

\usepackage{ACL2023}

\usepackage{times}
\usepackage{latexsym}

\usepackage[T1]{fontenc}

\usepackage[utf8]{inputenc}

\usepackage{microtype}

\usepackage{inconsolata}

\usepackage{multirow}
\usepackage{makecell}

\usepackage{times}
\usepackage{latexsym}
\usepackage{graphicx}
\usepackage{subfigure}
\usepackage{float}
\graphicspath{{images}}
\usepackage{stfloats}
\usepackage{booktabs}
\usepackage{indentfirst}
\usepackage{amsmath}
\usepackage{amssymb}

\usepackage{enumitem}

\usepackage{colortbl}
\usepackage{arydshln}

\usepackage[linesnumbered,ruled,vlined]{algorithm2e}

\SetCommentSty{mycommfont}

\SetKwInput{KwInput}{Input}                
\SetKwInput{KwOutput}{Output}              

\usepackage{CJKutf8}

\title{ALoRA: Allocating Low-Rank Adaptation for Fine-tuning \\ Large Language Models}

\author{
Zequan Liu$^1$ \ \ 
Jiawen Lyn$^2$ \  \
Wei Zhu$^3$\thanks{\ \ Corresponding author: wzhu@stu.ecnu.edu.cn.} \  \
Xing Tian$^4$  \ 
Yvette Graham$^2$\\ 
\textsuperscript{\rm 1} RWTH Aachen University, Aachen, Germany \\
\textsuperscript{\rm 2} Trinity College Dublin, Dublin, Ireland \\
\textsuperscript{\rm 3} East China Normal University, Shanghai, China \\
\textsuperscript{\rm 4} Niuxin Network Technology Co., Ltd. \\
}

\begin{document}
\maketitle
\begin{abstract}

Parameter-efficient fine-tuning (PEFT) is widely studied for its effectiveness and efficiency in the era of large language models. Low-rank adaptation (LoRA) has demonstrated commendable performance as a popular and representative method. However, it is implemented with a fixed intrinsic rank that might not be the ideal setting for the downstream tasks. Recognizing the need for more flexible downstream task adaptation, we extend the methodology of LoRA to an innovative approach we call allocating low-rank adaptation (ALoRA) that enables dynamic adjustments to the intrinsic rank during the adaptation process. First, we propose a novel method, AB-LoRA, that can effectively estimate the importance score of each LoRA rank. Second, guided by AB-LoRA, we gradually prune abundant and negatively impacting LoRA ranks and allocate the pruned LoRA budgets to important Transformer modules needing higher ranks. We have conducted experiments on various tasks, and the experimental results demonstrate that our ALoRA method can outperform the recent baselines with comparable tunable parameters.

\end{abstract}

\begin{CJK*}{UTF8}{gbsn}

\section{Introduction}

Large language models (LLMs) have been emerging and achieving state-of-the-art (SOTA) results not only on a variety of natural language processing tasks \cite{qin2023chatgpt,PromptCBLUE,text2dt_shared_task,Text2dt,zhu_etal_2021_paht,Li2023UnifiedDR,Zhu2023BADGESU,Zhang2023LECOIE,Zhu2023OverviewOT,guo-etal-2021-global,zhu-etal-2021-discovering,Zheng2023CandidateSF,info:doi/10.2196/17653,Zhang2023NAGNERAU,Zhang2023FastNERSU,Wang2023MultitaskEL,Zhu2019TheDS}, but also many challenging evaluation tasks \cite{huang2023c,li2023cmmlu,Cui2023UltraFeedbackBL} like question answering in special domains, reasoning, math, safety, instruction following. Despite LLMs becoming general task solvers, fine-tuning still plays a vital role in efficient LLM inference and controlling the style of the LLMs' generated contents.\footnote{Recently, OpenAI also released the fine-tuning API for GPT-3.5-turbo. See blog post: \url{https://openai.com/blog/gpt-3-5-turbo-fine-tuning-and-api-updates}.} Fine-tuning such large models by full parameters is prohibitive since it requires a large amount of GPU memory and computations. Thus, parameter-efficient fine-tuning (PEFT) \cite{Zhang2023LearnedAA,2023arXiv230318223Z,Ding2022DeltaTA} has raised much attention in the research field since in PEFT, the tunable parameters are often less than 1\% of the LLMs and the computation costs will be significantly decreased.

\begin{figure*}
\centering
\includegraphics[width=0.95\textwidth]{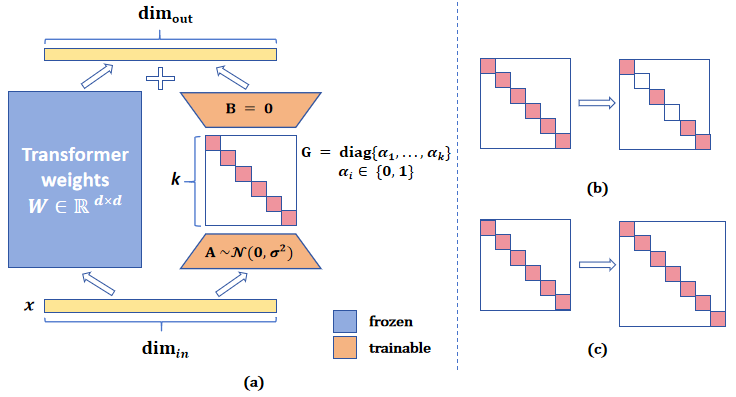}
\caption{Schematic illustration of our ALoRA. Left (a): ALoRA follows LoRA to update the weight matrix $W$ by fine-tuning the low-rank matrices $A$ and $B$ with intermediate rank $k$. Matrix $G$ is a diagonal matrix where each diagonal element is the gate unit $\alpha_{i}$ for each LoRA rank $i < k$. Each $\alpha_{i}$ is set to 1 at initialization. Right upper (b): Some abundant LoRA ranks are pruned by setting the corresponding gate $\alpha_{i}$ to zeros. Right lower (c): For weight matrix $W$ whose LoRA ranks are not pruned, we will assign additional LoRA ranks to enhance reparameterization. }
\label{fig:architecture}
\end{figure*}

Many PEFT methods have been validated to be effective across various models and tasks, often yielding comparable results with full-parameter fine-tuning \cite{He2021TowardsAU,zhu-tan-2023-spt,Zhang2023LearnedAA,Ding2022DeltaTA}. Among these PEFT methods, the reparameterization-based method low-rank adaptation (LoRA) \cite{hu2021lora} is considered one of the most efficient and effective methods at present. LoRA is especially popular after open-sourced LLMs become ubiquitous \cite{2023arXiv230514314D}. LoRA assumes that the change of the model’s parameters for adaptation is intrinsically low-dimensional and performs adaptation by optimizing the matrix obtained from low-rank decomposition. Since it is in the form of weight matrix reparameterization, LoRA parameters can be merged with the original LLMs and cause no forward propagation latency. 

Although LoRA is effective and can bring stable performance with the original setting in \citet{hu2021lora}, how to fully exploit its potential for downstream tasks still needs to be determined. First, how to determine the intrinsic rank for each model weight in the Transformer block is still unclear. Moreover, is it reasonable to set the same LoRA rank number for adapting the query, key, and value matrix? Second, in practice, the optimal LoRA rank setting would vary according to multiple factors, such as the backbone model and the task.

In order to improve the performance of downstream task adaptation of LoRA, we now propose the Allocating LoRA (ALoRA) framework (depicted in Figure \ref{fig:architecture}). First, LoRA modules with equal rank size are initialized at each Transformer weight, with all rank gates set to one. During fine-tuning, we re-allocate the LoRA ranks by (a) identifying which LoRA ranks are abundant or have negative contributions and prune those ranks by setting the rank gates to 0; (b) adding the pruned rank budgets to model weights that receive no pruning, that is, important model weights will be assigned more LoRA ranks. In order to calculate the contribution score of each LoRA rank efficiently and accurately, we propose a novel method, AB-LoRA. Our working procedure does not require re-training and does not require higher LoRA rank budgets at initialization or during training. 

We conduct extensive experiments on a wide collection of tasks, including sentiment classification, natural language inference, question answering, natural language generation under constraint, and instruction tuning, to demonstrate the effectiveness of our method. Notably, our method can consistently outperform strong PEFT baselines with comparable tunable parameter budgets, especially the recent LoRA variants. We also conducted an analysis showing that (a) our AB-LoRA method indeed can reflect the contribution of each LoRA rank; (b) our method can work with different LoRA rank budgets and different backbone models.

Our contributions are summarized as follows: 
\begin{itemize}
    \item We propose a novel method, AB-LoRA, to estimate the importance of each LoRA rank. 
    \item Built upon AB-LoRA, we propose our ALoRA framework, which can allocate LoRA ranks across different model weights and enhance the adaptation process. 
    \item We have conducted extensive experiments and analysis showing that our ALoRA framework is practical and outperforms the baselines under comparable parameter budgets. 
\end{itemize}

\section{Related works}

\subsection{Parameter-efficient fine-tuning (PEFT) methods}

Parameter-efficient fine-tuning (PEFT) is an approach of optimizing a small portion of parameters when fine-tuning a large pretrained backbone model and keeping the backbone model untouched for adaptation \cite{Ding2022DeltaTA,Zhang2023LearnedAA}. A branch of PEFT methods inserts additional neural modules or parameters into the backbone model. Representative works in this direction are Adapter \cite{houlsby2019parameter,Rckl2020AdapterDropOT,Zhang2023LearnedAA}, Prefix tuning \cite{li2021prefix}, Prompt tuning \cite{lester2021power}, P-tuning V2 \cite{Liu2022PTuningPT}. Another approach is to specify the particular parameters to be tunable or prunable \cite{BenZaken2021BitFitSP,guo-etal-2021-parameter,zhao-etal-2020-masking}. The reparameterization-based methods have attracted much attention \cite{hu2021lora}. This branch of approaches transforms the adaptive parameters during optimization into low-rank and parameter-efficient forms. This type of PEFT method is closely related to intrinsic dimension \cite{aghajanyan-etal-2021-intrinsic,2018arXiv180408838L}, that is, full parameter fine-tuning process of pre-trained models can be reparameterized into optimization within a low-dimensional subspace, i.e., fine-tuning has a low intrinsic dimension \cite{hu2021lora}. Intuitively, a well pretrained model does not need to be altered significantly for downstream task adaptation. \citet{Qin2021ExploringLI} investigate whether we can find a common intrinsic subspace shared by various NLP tasks. LoRA \cite{hu2021lora} is inspired by \cite{aghajanyan-etal-2021-intrinsic,2018arXiv180408838L} and hypothesizes that the change of weights during model tuning has a low intrinsic rank and optimizes the low-rank decomposition for the change of original weight matrices. PEFT methods are widely applied, especially with the popularization of open-sourced large language models \cite{2023arXiv230318223Z} and instruction tuning with these models for different application scenarios \cite{alpaca,2023arXiv230514314D}.

\subsection{The LoRA method and its variants}

LoRA \cite{hu2021lora} is proven to be effective and yield stable results when applied to both relatively small pretrained backbones and large language models \cite{2023arXiv230514314D,PromptCBLUE}. Despite its tractability and effectiveness, LoRA still has room for improvements in selecting optimal rank $r_{m}$ for each Transformer model weight $m$. The rank r takes discrete values; thus, changing it will directly alter the model structures. The optimal choices of ranks can vary across backbone models, tasks, and even Transformer model weights. Setting a large rank value for $r_{m}$ can waste training time and computation resources, while progressively setting a small $r_{m}$ may degrade the model performance. These limitations highlight the importance of upgrading LoRA with an adaptive strategy. 

There are already a few works investigating this direction. AdaLoRA \cite{Zhang2023AdaptiveBA} expresses the low-rank multiplication of LoRA in the form of singular value decomposition (SVD), and it identifies the most important ranks by a sensitivity-based importance score. SoRA \cite{Ding2023SparseLA} prunes abundant LoRA ranks by imposing a $l_{0}$ norm and optimizing with proximal gradient descent. SaLoRA \cite{Hu2023StructureAwareLA} prunes the LoRA ranks via the Lagrange multiplier method. Despite these recent efforts, we believe issues still need to be investigated for LoRA rank allocation: (a) The current works initialize a larger value for each $r_m$ and use certain heuristics to prune the number of ranks to meet a predefined budget. This training process inevitably requires additional GPU memory consumption. In addition, the maximum LoRA rank size for each model weight is limited, which restricts the solution space for LoRA rank allocations. (b) The current works depend on heuristic importance scores, which may not reliably reflect the contribution of each LoRA rank. Our work complements the existing literature by addressing the above issues.

\section{Methods}

\subsection{Preliminaries}

\noindent \textbf{Transformer model} \quad Currently, most widely used open-sourced language models and large language models adopt the stacked Transformer architecture \cite{Vaswani2017AttentionIA}. Each Transformer block is primarily constructed using two key submodules: a multi-head self-attention (MHA) layer and a fully connected feed-forward (FFN) layer. The MHA is given as follows:
\begin{equation}
x^{'} = MHA(xW^{Q}, xW^{k}, xW^{V})W^{O},
\end{equation}
where $MHA()$ denotes the multi-head attention operation,  $x \in \mathbf{R}^{l\times d}$ is the input tensor, $W^{O} \in \mathbf{R}^{d\times d}$ is the output projection layer (denoted as the Output module), and $W^{Q}, W^{K}, W^{V} \in \mathbf{R}^{d\times d}$ (denoted as the Query, Key, and Value modules). $l$ is the sequence length, $d$ is the hidden dimension. The FFN module consists of linear transformations and an activation function $g$ such as ReLU or GELU \cite{Hendrycks2016GaussianEL}. Take the FFN module in the LlaMA-2 models \cite{Touvron2023Llama2O} as example:
\begin{equation}
x^{'} = (g(xW^{G}) * xW^{U} ) W^{D}, 
\end{equation}
where $W^{G}, W^{U} \in \mathbf{R}^{d\times d^{'}}$ (denoted as Gate and Up module) and $W^{D} \in \mathbf{R}^{d^{'}\times d}$ (denoted as the Down module), and $d^{'}$ is usually larger than $d$. For notation convenience, we will refer to the number of modules in a Transformer block as $N_{mod}$. Thus, in LlaMA-2, $N_{mod} = 7$.

Denote the task's training set as $\mathcal{D}_{\text{train}} = {(x_m, y_m), m = 1, 2, ..., M}$, where $M$ represents the number of samples. In this work, we only consider the case where input $x_m$ and target $y_m$ are both text sequences. And we expect the language modeling head of LLMs to decode $y_m$ during inference. That is, no additional linear prediction heads are considered for predicting categorical or numerical values.

\subsection{Formulation}

Our objective is to efficiently fine-tune the LLMs for a specific downstream task under a given LoRA parameter budget $R^{target} = \sum_{m=1}^{N_{mod}} r_{m}^{target}$. The previous literature \cite{Ding2023SparseLA,Hu2023StructureAwareLA,Zhang2023AdaptiveBA} usually initialize the LoRA modules with a pre-defined large maximum rank $r_{max}$, consuming extra GPU memories. Different from the previous works, we now initialize each LoRA module with rank $r_{m}^{init} = R^{target} / N_{mod}$. That is, upon initialization, we have met the LoRA rank budget. Moreover, we will re-allocate the LoRA ranks in order to enhance the fine-tuning performance. 

In order to adjust the rank allocation of LoRA modules, we now inject gate units $\alpha_{i} \in \left\{0, 1\right\}$ ($i=1, 2, ..., r_{m}$) to each module $m$ with LoRA rank $r_{m}$. Imitating the formulation of SVD, the forward propagation of ALoRA is given by:
\begin{align}
z & = xW_{m}^{A}G_{m}W_{m}^{B}, \nonumber \\
G_{m} & = \text{diag}(\alpha_{m, 1}, ..., \alpha_{m, r_{m}}), 
\label{eq:lora_formulation}
\end{align}
where $\text{diag}()$ denotes a diagonal matrix, $W_{m}^{A} \in \mathbf{R}^{d\times r_m}$, $W_{m}^{B} \in \mathbf{R}^{r_m\times d}$. At initialization, the gate units are all set to 1.

Different from the previous literature \cite{Ding2023SparseLA,Hu2023StructureAwareLA,Zhang2023AdaptiveBA}, we take an alternative approach, that is, consider the problem of LoRA allocation as neural architecture search \cite{2023arXiv230108727W}. We consider the gate units $\alpha_{i}$ as architecture parameters (denoted as the set $\Theta$), the network with all the non-zero gate units as the super-network $M$, and denote the parameters in the down-projection and up-projection matrices as $\Omega$, then the optimization objective is:
\begin{align}
& \min_{\mathbf{\Theta}} \mathcal{L}(\mathcal{D}_{2}, \mathbf{\Omega}^{*}, \mathbf{\Theta}),  \nonumber\\
& \emph{s.t.} \ \ \mathbf{\Omega}^{*} =  \arg\min_{\mathbf{\Omega}} \mathcal{L}(\mathcal{D}_{1}, \mathbf{\Omega}, \mathbf{\Theta}), 
\label{eq:bi_level_optimize}
\end{align}
where $\mathcal{D}_{1}$ and $\mathcal{D}_{2}$ consists of a split of the training set $D_{train}$, $\mathcal{L}()$ is the loss function. This work uses the cross-entropy loss as the loss function. Note that with discrete values of $\alpha_{i}$, solving the above optimization problem is challenging due to non-differentiability. Thus, following the work of differentiable neural architecture search (DNAS) \cite{Liu2019DARTSDA}, $\alpha_{i}$ is relaxed to a continuous value in between (0, 1) and the equation \ref{eq:lora_formulation} becomes:
\begin{align}
z & = W_{m}^{B} G^{'}_{m} W_{m}^{A}x, \nonumber \\
G^{'}_{m} & = \text{diag}(\alpha^{'}_{m, 1}, ..., \alpha^{'}_{m,r_{m}}), \nonumber \\
\alpha^{'}_{m, i} & = 2 * \text{Sigmoid}(a^{'}_{m, i}), \text{ }a^{'}_{m, i} \in \mathbf{R},
\end{align}
where $a^{'}_{m, i}$ is initialized with zero value. With this setting, Equation \ref{eq:bi_level_optimize} becomes differentable and can be optimized by bi-level optimization \cite{Liu2019DARTSDA}.

\subsection{Our novel AB-LoRA method}

Under the DNAS setting, it is natural to consider the architecture weight $\alpha^{'}_{i}$ as the importance score for LoRA rank $i$, and one can use these scores to guide the pruning of abundant LoRA ranks. However, as pointed out by the literature \cite{Zhang2023LearnedAA,Chen2019ProgressiveDB}, and as will be demonstrated in the experiment, the architecture weights are not reliable indicators for the final LoRA allocation's performance. This observation motivates us to propose a simple yet effective modification to the DNAS-style architecture search. Instead of relying on the architecture weights' values to keep the best LoRA ranks, we propose directly evaluating the LoRA rank's superiority by its contribution or influence on the super-network's performances. Since our method mimics conducting ablation studies of a certain LoRA rank from the super-network, we refer to our method as the ablation-based LoRA (AB-LoRA).

\begin{algorithm}[!ht]
\DontPrintSemicolon
  
\KwInput{A super-network $M$, with $R^{target}$ LoRA ranks uniformly distributed in modules of $M$; }
\KwOutput{A new allocation of $R^{target}$ LoRA ranks. }
\KwData{Training set $D_{train}$, a batch of validation data $B_{val}$ }

Train super-network $M$ on the training set $D_{train}$ for $K_1$ epochs; 

\For{$n = 0$; $n < N_{A}$}
{
   \For{a single LoRA rank $r_{m, i}$ on $M$ }
   {
    Calculate the importance score $\text{IS}(r_{m, i})$ on $B_{val}$;
   }

    Prune $n_0$ LoRA ranks with lowest importance scores;

    \If{there are modules not pruned}
    {
        Add $n_0$ LoRA ranks to the un-pruned modules;
    }
    
    Further train the Super-network $M$ on $D_{train}$ for $K_2$ epochs; 
    
}
\caption{Workflow of ALoRA}
\label{algo:alora}

\end{algorithm}

We now introduce the core of our AB-LoRA method: calculating each LoRA rank's importance score, defined as how much it contributes to the performance of the super-network. Denote the complete super-network as $M$. Super-network $M$ is trained till convergence on the training set. We now consider a modified super-network obtained by zeroing out a single LoRA rank $r$ while keeping all other LoRA ranks. This new super-network is denoted as $M_{\backslash r}$. We also consider another modified super-network $M_{r}$ in which only LoRA rank $r$ is kept while all other LoRA ranks are zeroed out. We evaluate the three versions of super-networks on the same batch of validation data $B_{val}$. Denote the metric score as a function of a model $M$, $S(M)$, with the validation data fixed. Then, the importance score of LoRA rank $r$ is given by
\begin{equation}
\text{IS}(r) = S(M) - S(M_{\backslash r}) + S(M_{r}).
\label{eq:ab_lora}
\end{equation}
In the above equation, $S(M)$ can be treated as a constant term. Thus the above equation can be simplified to $\text{CS}(o) = - S(M_{\backslash r}) + S(M_{r})$. Intuitively, the LoRA rank that results in a significant performance drop upon zeroing out must play an important role in the super-network. Similarly, the one keeping most of the performance when acting alone contains important task-related knowledge and should be considered important. In the experiments, different from \citet{Chen2020StabilizingDA}, we set $S()$ as the negative of the cross-entropy (CE) loss since the widely applied metrics like accuracy or F1: (a) may not vary if the super-network only masks out a single operation, and (b) is not suitable for generative language model fine-tuning.

\subsection{The complete process of ALoRA}

With the guidance of the importance score in Equation \ref{eq:ab_lora}, we can now formally define the whole working process of our ALoRA framework (Figure \ref{fig:architecture}). Our working flow of allocating the LoRA ranks builds upon the following intuitions: (a) the pruning and allocation of LoRA ranks is conducted gradually to avoid performance degradation. (b) if the LoRA ranks in a Transformer module receive relatively high importance scores and are not pruned, this module is deemed important. It may need more LoRA ranks for adaptation so that the LoRA parameters can better learn the task knowledge.   

The framework of ALoRA is centered on our AB-LoRA method, which requires the super-network to be trained for $K_1$ epochs on the train set. We freeze the architectural parameters and train only the model parameters on the train set. No bi-level optimization is required, thus saving training time costs. Then, for each LoRA rank, we evaluate the importance score on a batch of samples $B_{val}$ from the development set. Then, $n_{A}$ LoRA ranks with the lowest scores are pruned by zeroing out their corresponding gate units. Moreover, if some Transformer modules do not have pruned LoRA ranks, we allocate the parameter budgets to them to enhance the adaptation further. \footnote{Note that increasing the rank size of a LoRA module from $r_{m}$ to $r_m^{'}$ for Transformer module $m$ involves concatenating newly initialized parameters for the matrices, so that $W_{m}^{A} \in \mathbf{R}^{d\times r_m}$ and $W_{m}^{B}\in \mathbf{R}^{r_m \times d}$ becomes $W_{m}^{A} \in \mathbf{R}^{d\times r_m^{'}}$ and $W_{m}^{B}\in \mathbf{R}^{r_m^{'} \times d}$. And the diagonal matrix $G_{m}$ is changed from $\text{diag}(\alpha_{m,1}, ..., \alpha_{m,r_m})$ to $\text{diag}(\alpha_{m,1}, ..., \alpha_{m,r_m^{m}})$. The newly added gate units are initialized with ones. }\footnote{If $n_{A}$ is not divided by the number of un-pruned modules, we allocate the $n_{A}$ ranks as uniformly as possible, with priority given to modules with higher average importance scores. For example, if $n_{A}=8$, and module $m_1$, $m_2$, $m_3$ are not pruned, and $m_1$ has the highest average importance score, $m_2$ ranks the second, $m_3$ receives the lowest average importance score. Then three ranks are given to $m_1$ and $m_2$, and two ranks are given to $m_3$.} After the pruning and adding operations, we tune the altered super-network for $K_2 > 0$ epochs to recover the lost performance. The above steps are repeated for $N_{A}$ times. Formally, we summarize the above process in Algorithm \ref{algo:alora}.

\section{Experiments}

In this section, we conduct a series of experiments to evaluate our ALoRA method.

\subsection{Baselines}

We compare our ALoRA framework with the current SOTA PEFT baseline methods. 

\noindent\textbf{Adapter-based tuning} \ We consider the following adapter tuning baselines: (1) Houlsby-Adapter \cite{houlsby2019parameter}; (2) Parallel-Adapter proposed by \citet{He2021TowardsAU}; (3) AdapterDrop \cite{Rckl2020AdapterDropOT}; (4) LST \cite{Sung2022LSTLS}; (5) Learned-Adapter \cite{Zhang2023LearnedAA}.

\noindent\textbf{Prompt-based tuning} \ For prompt-based tuning methods, we compare with (a) P-tuning v2 \cite{Liu2021PTuningVP}; (b) SPT \cite{zhu-tan-2023-spt}.

\noindent\textbf{LoRA and its variants} \ we consider the following LoRA variants as baselines: (a) LoRA \cite{hu2021lora}; (b) AdaLoRA \cite{Zhang2023AdaptiveBA}. (c) SoRA \cite{Ding2023SparseLA}; (d) SaLoRA \cite{Hu2023StructureAwareLA}. 

\noindent\textbf{Other PEFT methods} \ We also compare: (1) SSP \cite{Hu2022SparseSS}, which combines different PEFT methods.  

The baselines are implemented using their open-sourced codes. The hyper-parameter settings for the baselines are detailed in Appendix \ref{sec:appendix_exp_settings}.

\subsection{Datasets and evaluation metrics}

We compare our approach to the baselines on (a) four benchmark question-answering tasks: SQUAD \cite{rajpurkar-etal-2016-squad} and three tasks from the SuperGLUE benchmark\cite{Wang2019SuperGLUEAS} (BoolQ, COPA and ReCoRD). (b) three sentence level tasks from GLUE benchmark \cite{Wang2018GLUEAM}, SST-2, RTE, QNLI. (d) Alpaca dataset \cite{alpaca} for instruction tuning, and MT-Bench \cite{2023arXiv230605685Z}, to evaluate the instruction tuning quality of LLMs. The dataset introductions, statistics, and prompt-response templates for the above tasks are detailed in Appendix \ref{sec:appendix_datasets}. The above tasks' evaluation metrics or protocols are in Appendix \ref{sec:appendix_evaluations}.

\begin{table*}[tb!]
\centering
\resizebox{0.88\textwidth}{!}{
\begin{tabular}{c|cc|ccccccc}
\hline
\multirow{2}*{\textbf{Method}}   &   \multicolumn{2}{c}{\textbf{Additional Params}}    &     \textbf{SST-2}   &    \textbf{RTE}   &   \textbf{QNLI}   &   \textbf{BoolQ}  &  \textbf{COPA}    &   \textbf{ReCoRD}   &    \textbf{Squad}  \\ 

&  \textbf{Initial}  &  \textbf{Final}   &  \textbf{(acc)}   &   \textbf{(acc)}     &  \textbf{(acc)}   &   \textbf{(acc)}  &   \textbf{(acc)}  &   \textbf{(f1-em)}   &   \textbf{(f1-em)}  \\
\hline

\multicolumn{10}{c}{\textbf{\emph{Baselines}}}  \\
\hline

P-tuning v2    &   20.9M  &    20.9M    &   93.4    &   79.6  &   92.6   &   84.7   &   90.3    &  89.9   &   87.6    \\
SPT  &   16.8M  &   16.8M   &    93.6   &  80.3    &   92.8   &  85.3     &   90.6   &  90.2    &   88.1  \\
\hdashline

Housbly-Adapter   &  21.0M  &   21.0M    &   93.5  &   81.3   &  92.9    &  85.2   & 91.0    &  90.4   &  88.0  \\
Parallel-Adapters  &  21.0M  &   21.0M   &    93.6   &  81.2   &  93.0   &  85.7    & 90.8    & 90.6    &  88.2   \\
AdapterDrop   &   21.0M  &   21.0M    &    93.2   &  80.7  &  92.8   &   85.1    &   90.6   &  90.3   &  87.9  \\
LST    &  21.1M    & 21.1M     &   93.4    &  81.6   &  93.0   &  86.2   &  91.0   &  90.4    &  87.9   \\
Learned-Adapter   &   21.2M   &  21.2M    &  94.1    &  82.1  &  93.1   &   87.0   &  91.1   &  90.7   &   88.3   \\
\hdashline


LoRA   &  20.0M   &  20.0M    &   94.1    &   83.3   &  93.1    &  87.3   &  91.3   &  90.8   &  88.4  \\
AdaLoRA   &   40.0M   &   20.0M   &  94.1    &  83.5  &  93.2   &  87.1   &   91.6    &  \underline{91.1}   &  88.3  \\
SoRA    &   40.0M   &   20.0M  &  \underline{94.2}   &     \underline{83.7}   &  \underline{93.3}   &  \underline{87.6}   & \underline{91.7}    &  91.0   &  \underline{88.5} \\
SaLoRA   &   40.0M   &   20.0M   &    93.9   &   83.4  &  93.2   & 
 87.2    &  91.5    &  90.9   &  88.4  \\

\hdashline
SSP &   40.0M  &  20.0M   &   94.1   &  83.1  &  93.1    &  87.1   &  91.6    &  90.6   &  88.2   \\

\hline
\multicolumn{10}{c}{\textbf{\emph{Our proposed methods}}}  \\
\hline

ALoRA   &   20.0M   &  19.6M   &   \textbf{95.0}   &   \textbf{84.6}    &  \textbf{93.7}  &   \textbf{88.0}     &   \textbf{92.1}   &   \textbf{91.8}   &   \textbf{89.2}    \\

\hline
\end{tabular}}

\caption{\label{tab:results_main_1} The Overall comparison of the three GLUE tasks and four question-answering tasks. The backbone model is LlaMA-2 7B. We report the median performance over five random seeds. Bold and Underline indicate the best and the second-best results. The metric for each task is explained in Appendix \ref{sec:appendix_evaluations}.} 

\end{table*}

\subsection{Experiment Settings}

\noindent\textbf{Computing infrastures} \quad We run all our experiments on NVIDIA A40 (48GB) GPUs. 

\noindent\textbf{Pretrained backbones} \quad The main experiments uses most recent open-sourced LLM, LlaMA-2 7B released by Meta \cite{Touvron2023Llama2O} as the pretrained backbone model. In the ablation studies, we will also use GPT2-large model \cite{radford2019language}, and RoBERTa-large \cite{Liu2019RoBERTaAR}.

\noindent\textbf{Prediction heads} \quad When fine-tuning LlaMA-2 7B, we only consider the supervised fine-tuning (SFT) setting \cite{ouyang2022training}, that is, all the predictions are generated using the language modeling head (LM head) upon receiving a prompt or an instruction. For decoding during inference, we use beam search with beam size 5.

\noindent\textbf{Hyper-parameters for ALoRA} \quad In our experiments, unless otherwise specified, we set $R^{target}$ to $8 * N_{mod}$, and initially all Transformer model weights are paired with LoRA modules with rank $r_{m}^{init} = 8$. In this setting, ALoRA satisfies the LoRA rank budget upon initialization, and thus during training and inference.\footnote{Note that it is possible that the total LoRA ranks after training is smaller than that at initialization. } We set $n_{A}$ to $1 * N_{mod}$. For training with the ALoRA's workflow, we set the batch size of $B_{val}$ to 32, $K_1$ to 1 epoch, $K_2$ to 0.25 epoch, and the LoRA rank allocation procedure is conducted for at most $N_{A} = 8$ times.

\noindent\textbf{Reproducibility} \quad We run each task under five different random seeds and report the median performance on the test set of each task. 

Due to limited length, other experimental settings for the baseline methods and the training procedure are put in Appendix \ref{sec:appendix_exp_settings}.

\subsection{Main results}
\label{subsec:main_results}

The experimental results on the three classification tasks and 4 question answering tasks are presented in Table \ref{tab:results_main_1}. In the second and third columns of Table \ref{tab:results_main_1}, we present the initial number of tunable parameters and the final ones. Table \ref{tab:results_main_1} reveals that our ALoRA method outperforms the baseline methods across all seven tasks, with comparable or fewer tunable parameters throughout the training and inference processes. In particular, ALoRA successfully outperforms AdaLoRA, SoRA, and SaLoRA with comparable initial and final LoRA parameters. These results demonstrate that our method can better allocate LoRA parameters for better downstream task adaptation.

For the E2E benchmark \cite{novikova-etal-2017-e2e}, the results are reported in Table \ref{tab:results_e2e}. The results show that on the E2E task, our ALoRA method successfully outperforms LoRA and SoRA regarding BLEU, ROUGE-L, or METEOR scores.

\begin{table}[tb!]
\centering
\resizebox{0.47\textwidth}{!}{
\begin{tabular}{c|ccc}
\hline
\textbf{Method}   &    \textbf{BLEU}   &   \textbf{ ROUGE-L}   &   \textbf{METEOR}    \\ 
\hline

Learned-Adapter   &  68.9    &    70.9    &   45.8   \\
LoRA   &     68.9   &  \underline{71.2}   &  46.1     \\
SoRA   &   \underline{70.0}   &   71.1   &    \underline{46.3}   \\
\hdashline
ALoRA   &   \textbf{70.6}   &  \textbf{71.8}  &     \textbf{47.1}     \\

\hline
\end{tabular}}
\caption{\label{tab:results_e2e} Results for different PEFT methods on the E2E benchmark. The backbone LM is LlaMA-2 7B. The metrics are explained in Appendix \ref{sec:appendix_evaluations}.}
\end{table}

\begin{table}[tb!]
\centering
\resizebox{0.48\textwidth}{!}{
\begin{tabular}{c|cc}
\hline
\textbf{Method}   &    \textbf{Avg GPT-4 score} ($\uparrow$)   &    \textbf{ ROUGE-L} ($\uparrow$)    \\ 
\hline
SoRA    &   7.16   &    53.2 \\
\hdashline
ALoRA   &   7.47   &   54.3  \\
\hline

\end{tabular}}
\caption{\label{tab:results_alpaca} The performance of instruction tuning using the SoRA and ALoRA methods. The backbone model is LlaMA-2 7B. $\uparrow$ means the metric is higher the better. }

\end{table}

\begin{figure*}[ht]	
\centering
\subfigure[BoolQ]{%
\includegraphics[width=0.37\textwidth]{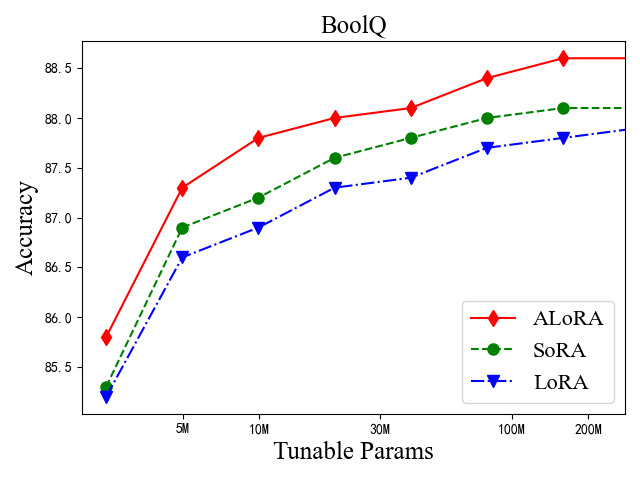}
\label{subfig:different_rank_boolq}
}
\subfigure[E2E]{%
\includegraphics[width=0.37\textwidth]{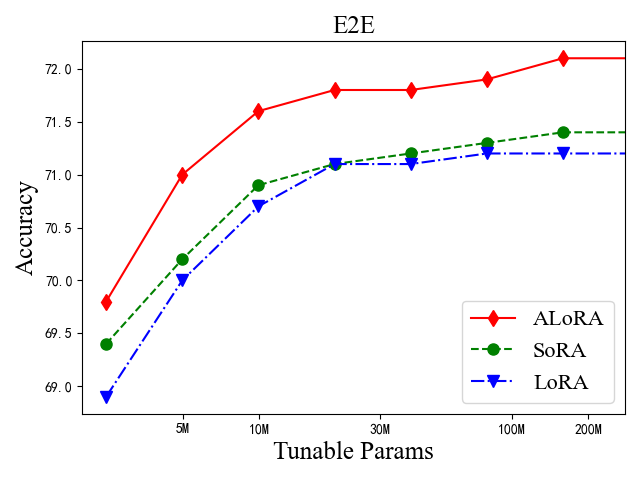}
\label{subfig:different_rank_e2e}
}
\caption{Performances under different LoRA rank budgets. The $x$-axis represents the number of tunable parameters, and the $y$-axis represents the performance score. }
\label{fig:different_rank_value}

\end{figure*}

After the LlaMA-2 7B is fine-tuned on the Alpaca dataset with our ALoRA and SoRA methods, we utilize the 80 instructions in the MT-Bench as the test set. We follow the current standard practice of utilizing GPT-4 as an unbiased reviewer \cite{2023arXiv230605685Z}. The protocol of utilizing GPT-4 as the reviewer and scorer is specified in Appendix \ref{sec:appendix_evaluations}. The average score provided by GPT-4 is presented in Table \ref{tab:results_alpaca}, along with the ROUGE-L scores calculated by considering the GPT-4's answers as ground truth. Consistent with the previous experiments (Table \ref{tab:results_main_1} and \ref{tab:results_e2e}), our ALoRA method outperforms the SoRA method in terms of the GPT-4 evaluation scores and ROUGE-L, demonstrating that ALoRA can enhance the instruction tuning quality of large language models. A case study of answers generated by different methods is presented in Table \ref{tab:results_alpaca_examples_1}, showcasing that ALoRA leads to better instruction-tuned LLMs.

\subsection{Ablation studies and analysis}

\begin{table}[tb!]
\centering
\resizebox{0.48\textwidth}{!}{
\begin{tabular}{c|ccc}
\hline
\textbf{Method}   &    \textbf{Memory cost (GB)}  &  \textbf{Speed (it/s)}   &   \textbf{Time cost (h)}     \\ 
\hline

LoRA   &   17.6   &   5.01   &   2.68  \\
SoRA    &   18.8   &    4.96  &  3.63   \\
\hdashline
ALoRA   &   18.1   &   5.01   &   3.81 \\
\hline

\end{tabular}}
\caption{\label{tab:results_efficiency_analysis} The memory, speed and time cost for fine-tuning LlaMA-2 7B on the E2E task with different PEFT methods. }
\end{table}

\noindent\textbf{Analysis of Training Efficiency} \quad So far, we have demonstrated that our ALoRA can outperform LoRA and SoRA on a wide collection of tasks. One might suspect this advantage is achieved with significant time or memory costs. We compare the max training GPU memory, training speed, and training time costs of ALoRA, SoRA, and LoRA when fine-tuning LlaMA-2 7B with the E2E benchmark. From Table \ref{tab:results_efficiency_analysis}, one can see that ALoRA requires less memory costs during training than SoRA since it does not initialize with a larger LoRA rank. Moreover, under early stopping, the total training time cost of ALoRA remains comparable with SoRA and LoRA.

\noindent\textbf{Ablation study of ALoRA framework} \quad We now consider the following variants of ALoRA: (a) instead of utilizing our novel AB-LoRA method, we follow the optimization procedure of Equation \ref{eq:bi_level_optimize}, and use the architectural weights $\alpha_{m, i}^{'}$ as the importance scores. This variant is denoted as ALoRA-DNAS. (b) Use the sensitivity-based metric in \citet{Zhang2023AdaptiveBA} as the importance measurement. (denoted as ALoRA-Sensi). The experimental results on the BoolQ, ReCoRD, and SQUAD tasks are reported in Table \ref{tab:results_ablation_alora} of Appendix \ref{sec:appendix_ablation_alora}. The results show that ALoRA outperforms the two variants, demonstrating that our AB-LoRA method can provide better guidance in allocating LoRA ranks.

\begin{figure}
\centering
\includegraphics[width=0.48\textwidth]{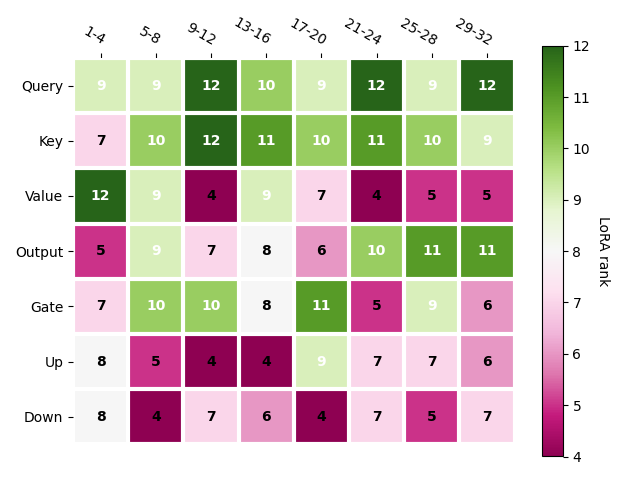}
\caption{The final rank allocations of ALoRA after fine-tuning the LlaMA-2 7B model on the E2E task.}
\label{fig:visualize}
\end{figure}

\noindent\textbf{Visualization of the final rank allocations} \quad In this section, we visualize the final rank allocations of ALoRA after the training process on the E2E task in Figure \ref{fig:visualize}. We also compare the LoRA rank allocations by the SoRA method in Figure \ref{fig:visualize_sora} of Appendix \ref{sec:appendix_visualize}. We can see from Figure \ref{fig:visualize} that (a) More LoRA rank budgets are put to adapt the query and key modules, while the value and output modules in the self-attention are less emphasized. (b) The feed-forward layer in the Transformer block requires fewer LoRA ranks, indicating that this layer stores general language knowledge, while the attention module will contain more task-specific knowledge after ALoRA fine-tuning. Compared with ALoRA's allocation, SoRA results in a more unbalanced allocation, putting more rank budgets to the Down module than our ALoRA method.

\noindent\textbf{Comparisons under different LoRA rank budgets} Note that in the main experiments, we set the targeted LoRA rank budget as $R^{target} = 8 * N_{mod}$. Now we vary this budget to any multiplier in {1, 2, 4, 8, 16, 32, 64, 128} times $N_{mod}$, and see how ALoRA, SoRA, and LoRA perform on the BoolQ and E2E tasks. The experimental results are presented in Figure \ref{subfig:different_rank_boolq} and \ref{subfig:different_rank_e2e}. From the results, we can see that under different LoRA rank budgets, our ALoRA method can consistently outperform LoRA and SoRA by effectively allocating different LoRA ranks properly to different Transformer modules, thus enhancing the performance of fine-tuning.

\noindent\textbf{Ablation on the pretrained backbones} \quad Our main experiments are conducted on the LlaMA-2 7B model. To demonstrate the wide applicability of our method, we now conduct experiments on RoBERTa-large and GPT2-large. The results are reported in Table \ref{tab:results_roberta_large} and \ref{tab:results_e2e_gpt2}. We can see that on these two backbones, our method can also outperform the baseline methods.

\section{Conclusion}

This work presents the Allocating Low-Rank Adaptation (ALoRA), an innovative method for parameter-efficient fine-tuning large language models. Upon the hypothesis that the adaptation for different Transformer modules could be of different tanks, we introduce a novel workflow for allocating LoRA ranks in the fine-tuning process. First, we propose a novel method, AB-DNAS, to accurately evaluate the importance scores of LoRA ranks. Second, guided by the AB-DNAS method, our workflow allows the pruning of ranks at specific modules and considers allocating more ranks to essential modules. Thus, our method does not require to set a more significant initial rank. Our method is convenient to implement and off-the-shelf. Experiments on various tasks demonstrate that our ALoRA method outperforms the baseline methods.

\section{Acknowledgements}

Jiawen Lyn and Yvette Graham's contribution was conducted with the financial support of the Science Foundation Ireland Centre for Research Training in Digitally-Enhanced Reality (d-real) under Grant No. 18/CRT/6224 and ADAPT at Trinity College Dublin
under Grant Agreement No 13/RC/2106\_P2.  For the purpose of Open Access, the author has applied a CC BY public copyright licence to any Author Accepted Manuscript version arising from this submission.

\section*{Limitations}

We showed that our proposed method can greatly improve the performance of parameter-efficient tuning on diverse tasks and different pretrained models (i.e., LlaMA-2 7B, RoBERTa-large and GPT2-large). However, we acknowledge the following limitations: (a) the more super-sized open-sourced LLMs, such as LlaMA-2 13B and 70B, are not experimented due to limited computation resources. (b) Other tasks in natural language processing, like information extraction, were also not considered. But our framework can be easily transferred to other backbone architectures and different types of tasks. It would be of interest to investigate if the superiority of our method holds for other large-scaled backbone models and other types of tasks. And we will explore it in future work.

\section*{Ethics Statement}

The finding and proposed method aims to improve the low-rank adaptation (LoRA) based tuning in terms of better rank allocations and performances. The used datasets are widely used in previous work and, to our knowledge, do not have any attached privacy or ethical issues. In this work, we have experimented with LlaMA-2 7B, a modern large language model. As with all LLMs, LlaMA-2’s potential outputs cannot be predicted in advance, and the model may in some instances produce inaccurate, biased or other objectionable responses to user prompts. However, this work's intent is to conduct research on different fine-tuning methods for LLMs, not building applications to general users. In the future, we would like to conduct further testing to see how our method affects the safety aspects of LLMs.

\bibliography{custom}
\bibliographystyle{acl_natbib}

\appendix

\section{Additional related works}

\noindent\textbf{Adapter-based tuning.} \quad One of the most important research lines of PEFT is adapter-based tuning. Adapter \cite{houlsby2019parameter} inserts adapter modules with bottleneck architecture between every consecutive Transformer \cite{Vaswani2017AttentionIA} sublayers. AdapterFusion \cite{pfeiffer-etal-2021-adapterfusion} only inserts sequential adapters after the feed-forward module. Adapter-based tuning methods have comparable results with model tuning when only tuning a fraction of the backbone model's parameter number. Due to their strong performance, a branch of literature has investigated the architecture of adapters in search of further improvements. \citet{He2021TowardsAU} analyze a wide range of PETuning methods and show that they are essentially equivalent. They also propose the general architecture of PEFT, and derive the Parallel Adapter which connects the adapter modules in parallel to the self-attention and MLP modules in the Transformer block. AdapterDrop \cite{Rckl2020AdapterDropOT} investigates the efficiency of removing adapters from lower layers. Adaptive adapters \cite{Moosavi2022AdaptableA} investigate the activation functions of adapters and propose to learn the activation functions of adapters via optimizing the parameters of rational functions as a part of the model parameters. Compacter \cite{Mahabadi2021CompacterEL} uses low-rank parameterized hypercomplex multiplication \cite{Le2021ParameterizedHG} to compress adapters' tunable parameters. LST \cite{Sung2022LSTLS} improves the memory efficiency by forming the adapters as a ladder along stacked Transformer blocks, and it enhances the adapter module by adding a self-attention module to its bottleneck architecture. \cite{Sung2022LSTLS,Jie2022ConvolutionalBA} try to add different encoding operations, like self-attention operations and convolutions between the bottleneck structure of adapters, and achieve better performances. Learned-Adapter \cite{Zhang2023LearnedAA} builds upon the above adapter-based methods and enhance the performance of adapter tuning by automatically learning better architectures for adapters.

\noindent\textbf{Prompt tuning methods} \quad Prompt tuning \cite{lester2021power} and P-tuning \cite{Liu2022PTuningPT} insert a soft prompt to word embeddings only, and can achieve competitive results when applied to supersized PTMs. Prefix-tuning \cite{li2021prefix} and P-tuning v2 \cite{Liu2021PTuningVP} insert prompts to every hidden layer of PTM. IDPG \cite{Wu2022IDPGAI} uses the prompt generator with parameterized hypercomplex multiplication \cite{Le2021ParameterizedHG} to generate a soft prompt for every instance. LPT \cite{Liu2022LatePT} improves upon IDPG by selecting an intermediate layer to start inserting prompts. SPT \cite{zhu-tan-2023-spt} designs a mechanism to automatically decide which layers to insert new instance-aware soft prompts.

\noindent\textbf{LoRA methods} \quad Since LoRA is the most popular PEFT method in the era of large language models, there are many works that are orthogonal to AdaLoRA, SoRA and our work that are devoted to improve LoRA on many different aspects. QLoRA \cite{2023arXiv230514314D} proposes a novel quantization method that can significantly reduce the memory consumptions of LLMs during LoRA fine-tuning. LoRA-FA \cite{Zhang2023LoRAFAML} freezes parts of the randomly initialized LoRA matrices. (d) VERA \cite{Kopiczko2023VeRAVR} investigate whether one could froze the randomly initialized LoRA matrices and only learns a set of scaling vectors. Tying LoRA matrices across layers are also investigated by VERA.

\section{Appendix for the datsets and evaluation metrics}
\label{sec:appendix_datasets}

\subsection{Datasets from GLUE and SuperGLUE }

We experiment on three tasks from the GLUE \cite{Wang2018GLUEAM} benchmark: (a) (a) a sentiment classification task, SST-2. (b) two benchmark natural language inference tasks, RTE and QNLI. We also experiment with three question-answering tasks: (a) two question answering tasks in the format of binary choices, COPA and BoolQ. (b) A Squad \cite{rajpurkar-etal-2016-squad} style question answering task, ReCoRD.

Since the original test sets are not publicly available for these tasks, we follow \citet{Zhang2020RevisitingFB,Mahabadi2021CompacterEL,ACF,Gao2023FPABEEFE,Zhu2021MVPBERTMP,zhu-2021-leebert,zhang-etal-2022-pcee,zuo-etal-2022-continually,sun-etal-2022-simple,zhu-etal-2021-gaml,zhu2021autonlu,autotrans,li-etal-2019-pingan,zhu-etal-2019-panlp,zhu-2021-autorc,zhang2021automatic,Wang2020MiningIH} to construct the train/dev/test splits as follows to ensure a fiar comparison: (a) for datasets with fewer than 10k samples (RTE, COPA, BoolQ), we divide the original validation set in half, using one half for validation and the other for testing. (b) for larger datasets, we split 1k samples from the training set as the development set, and use the original development set as the test set. The detailed statistics of the GLUE and SuperGLUE benchmark tasks is presented in Table \ref{tab:dataset_stats}.

\begin{table*}[tb!]
\centering
\resizebox{1.0\textwidth}{!}{
\begin{tabular}{cccccccc}
\hline
Datasets  &  \#train    &  \#dev   &   \#test   &   $ | \mathcal{Y} | $   &   Type   &  Labels  &  Metrics  \\ 
\hline
\multicolumn{8}{c}{\textbf{\emph{SuperGLUE tasks}}}  \\
\hline
BoolQ  &  9.4k   &    1.6k   &  1.6k   &    2   &   Question Answering    &  True, False   &   acc \\
COPA   &   0.4k   &   0.05k   &   0.05k    &   2   &      Question Answering   &    choice1, choice2   &   acc   \\
ReCoRD  &    101k &    1k   &    7.4k  &  - 
 &  Question Answering   &  -   &  f1-em  \\
\hline
\multicolumn{8}{c}{\textbf{\emph{GLUE tasks}}}  \\
\hline
SST-2  &  66k  &   1k    &   0.8k   &   2   &  sentiment classification &  positive, negative    &  acc     \\
RTE &   2.5k   &   0.1k   &    0.1k  &    2   &    NLI   &    entailment, not entailment  &  acc    \\
QNLI &  104k   &    1k   &    5.4k  &   2   &   NLI   &   entailment, not entailment   &  acc  \\

\hline
\multicolumn{8}{c}{\textbf{\emph{Other tasks}}}  \\
\hline

Squad &   87k  &  1k    &   5.9k   &   -   &  Question Answering 
 &  -   &   f1-em   \\
E2E &   42k  &  4.6k   &  4.6k   &   -   &  NLG   &  -    &  BLEU/ROUGE-L/METEOR   \\
Alpaca   &    52k  &  -   &  -   &  -  &  Instruction tuning  &  -   &   -   \\
MT-Bench   &   -   &  -  &   80  &   -   &    Instruction tuning  &  -    &  GPT-4 scores       \\

\hline
\end{tabular}}
\caption{\label{tab:dataset_stats}  The dataset statistics of the GLUE and SuperGLUE benchmark tasks evaluated in this work. $ | \mathcal{Y} | $ is the number of classes for a classification task. }
\end{table*}

\subsection{The Squad task}

Stanford Question Answering Dataset (SQuAD) \cite{rajpurkar-etal-2016-squad} is a reading comprehension dataset, consisting of questions posed by crowdworkers on a set of Wikipedia articles, where the answer to every question is a segment of text, or span, from the corresponding reading passage, or the question might be unanswerable. This task is one of the most widely studied question answering task in the field. 

In this work, we use the v1.1 version of SQUAD. Since the original test sets are not publicly available for these tasks, we follow \citet{Zhang2020RevisitingFB,Mahabadi2021CompacterEL} and split 1k samples from the training set as the development set, and use the original development set as the test set. The detailed statistics of this task is presented in Table \ref{tab:dataset_stats}.

\subsection{Datasets: E2E benchmark}

The E2E benchmark dataset for training end-to-end, data-driven natural language generation systems in the restaurant domain. It asks a model to generate natural utterances based on a set of given key contents. This dataset has a 42061/4672/4693 train/dev/test split.

\subsection{Dataset: Instruction tuning}

Instruction tuning is an important method to improve the general capabilities of large language models \cite{ouyang2022training}. With the rise of large language models in the scale of 10B parameters or more,  like GPT-3, T5, PaLM, researchers have actively explored the few-shot or zero-shot capabilities of these models. \cite{Mishra2021CrossTaskGV} find that fine-tuning these LLMs on a large scale datasets containing hundreds of NLP tasks significantly improves the zero-shot performances on unseen tasks, establishing the scaling law of task numbers. The previous works like \cite{Wei2021FinetunedLM} and T0 \cite{Sanh2021MultitaskPT} establishes the instruction tuning datasets by transforming the traditional NLP tasks into a unified prompt format. Instruct-GPT \cite{ouyang2022training} conducts instruction tuning using the dataset constructed based the user queries from the OpenAI API users. Note that this work is also a seminal work for human feedback learning with reinforcement learning. However, the complete instruction tuning dataset from \cite{ouyang2022training} remains closed. With the launch of ChatGPT, \cite{alpaca} (Alpaca) constructs an instruction tuning dataset with diverse topics using the self-instruct techniques. 

 For our experiment, we employ the Alpaca dataset \cite{alpaca} for instruction tuning. Specifically, we employs its cleaned version\footnote{\url{https://huggingface.co/datasets/yahma/alpaca-cleaned}.}. This dataset comprises 51K instructions and demonstrations, and is suitable for instruction tuning. The cleaned version corrects multiple issues such as hallucinations, merged instructions, and empty outputs.

\subsection{Evaluation metrics/protocols}
\label{sec:appendix_evaluations}

For the three GLUE tasks we experiment on, we report accuracy (denoted as acc). For ReCoRD, we report the average of the F1 score and the exact match score (denoted as f1-em). For the BoolQ and COPA tasks, we report accuracy. The above choices of evaluation metrics strictly follow \cite{Wang2018GLUEAM} and \cite{Wang2019SuperGLUEAS}. 

For the SQUAD dataset, we also report the average of the F1 score and the exact match score (denoted as f1-em).

Following \cite{novikova-etal-2017-e2e}, we report three different metrics on the E2E task: (a) BLEU; (b) ROUGE-L; (c) METEOR. We rely on the HuggingFace Evaluate package\footnote{\url{https://huggingface.co/docs/evaluate/index}} for computing these metrics.

For evaluating the quality of instruction tuned LlaMA-2 7B, we follow the current common practice of utilizing GPT-4 as a unbiased reviewer \cite{2023arXiv230605685Z}. 80 instructions from the MT-Bench is set as a test set. We generate model responses from a fine-tuned model with beam size 5 with the generation function in Huggingface Transformers \cite{wolf2020transformers}. Then we compare SoRA and ALoRA's answers with GPT-4. For each instruction in MT-Bench, GPT-4 \cite{gpt4} is asked to write a review for both answers from the two methods, and assigns a quantitative score on a scale of 10 to each response. The prompts of instructing GPT-4 for evaluation is presented in Appendix \ref{sec:appendix_gpt4_eval}. ROUGE-L scores computed by considering the answers generated by GPT-4 as the ground truth.

\section{Prompt templates for fine-tuning LlaMA-2 7B}
\label{sec:appendix_prompt_templates}

Since we fine-tune LlaMA-2 7B without introducing task-specific prediction heads, we need to transform all the tasks into a prompt-response format. Now we present the prompt-response template for each task. 

\noindent \textbf{Templates for RTE and QNLI} Since these two tasks are NLI tasks, the samples in them consists of two input text, [sentence1] and [sentence1], and a label [label\_name] (entailment or not entailment). Thus, we use the following templates:

Template for prompt: 
\begin{verbatim} 
sentence 1: [sentence1]
sentence 2: [sentence1]
Are sentence 1 and sentence 2 have 
entailment relation or not?
\end{verbatim}

Template for response:
\begin{verbatim} 
[label_name]
\end{verbatim}

\noindent \textbf{Templates for SST-2} The samples in this task consists of one input text, [sentence], and a label [label\_name] (positive or negative). 

Template for prompt: 
\begin{verbatim} 
[sentence] 
The sentiment of the given sentence is:
\end{verbatim}

Template for response:
\begin{verbatim} 
[label_name]
\end{verbatim}

\noindent \textbf{Templates for BoolQ} The samples in this task consists of a reference document, [doc], a query, [query], and a label [label\_name] (yes or no). 

Template for prompt: 
\begin{verbatim} 
Reference document:
[doc]
Question:
[query]
\end{verbatim}

Template for response:
\begin{verbatim} 
[label_name]
\end{verbatim}

\noindent \textbf{Templates for COPA} The samples in this task consists of a premise, [premise], two choices, [choice1] and [choice2], a query, [query], and a label [label\_name] (1 or 2, indicating which choice is consistent with the premise). 

Template for prompt: 
\begin{verbatim} 
Premise:
[premise]
Choice 1: [choice1]
Choice 2: [choice2]
Question:
[query]
\end{verbatim}

Template for response:
\begin{verbatim} 
[label_name]
\end{verbatim}

\noindent \textbf{Templates for ReCoRD and SQUAD} The samples in these two tasks consist of a context document, [context], a question, [query], and a answering span, [answer]. 

Template for prompt: 
\begin{verbatim} 
Context:
[context]
Question:
[query]
\end{verbatim}

Template for response:
\begin{verbatim} 
[answer]
\end{verbatim}

\noindent \textbf{Templates for E2E} The samples in this task consists of a reference [ref], consisting required information, and a targeted response, [target], which is a customer review written according to the reference's contents. 

Template for prompt: 
\begin{verbatim} 
Reference:
[ref]
Generate a customer review following the 
given reference.
\end{verbatim}

Template for response:
\begin{verbatim} 
[target]
\end{verbatim}

\section{Prompt templates for GPT-4 evaluations}
\label{sec:appendix_gpt4_eval}
In this work, we utilize the powerful LLM GPT-4 \cite{gpt4} as the evaluator for comparing the instruction tuning quality. As a reviewer, GPT-4 will receive a query [query], two responses, [response1] and [response2], from two assistants. We will ask GPT-4 to write a review for each response, assessing the quality of the response, and then ask GPT-4 to assign a score on a scale of 10 to each response.

Template for prompt: 
\begin{verbatim} 
Task Introduction
you will be given a query, and two responses 
from two assistants, 
could you compare the two responses, 
and do the following: 
(1) write a concise review for each 
assistant's response, on how well the 
response answers the query, and whether 
it will be helpful to humans users, and any 
issues in the response;
(2) assigns a quantitative score on a scale 
of 10 to each response, reflecting 
your assessment of the two responses
Query: 
[query]
Response 1 from assistant 1: 
[response1]
Response 2 from assistant 2: 
[response2]
\end{verbatim}

\section{Appendix for Experimental settings}
\label{sec:appendix_exp_settings}

Here, we provide more details for experimental settings. 

\noindent\textbf{Hyper-parameters for the baseline PEFT methods} \quad For P-tuning V2, the number of prompt tokens at each layer is set to 160. For SPT, the bottleneck dimension is set to 256, and the number of prompt layers is set to 8. For adapter-based methods, the bottleneck dimension is set to 40, and the adapter modules are added on the self-attention and feed-forward module. For LoRA and ALoRA, the initial rank at each module is set to 8. For AdaLoRA, SoRA, and SaLoRA, the initial rank at each module is set to 16, and half of the rank budget is pruned during fine-tuning. We adjust the sparsity for SSP so that the number of tunable parameters is comparable with ALoRA and the other baselines.

\noindent\textbf{Training settings for PEFT methods} \quad We use the HugginFace Transformers \cite{wolf-etal-2020-transformers} and PEFT \cite{peft} for implementing all the methods, and for training and making predictions. For fine-tuning LlaMA-2 7B model, the maximum sequence length is set to 2048. The maximum training epoch is set to 10. The batch size is set between 16 for task with less than 10k training set, and 128 otherwise. We use AdamW as the optimizer with a linear learning rate decay schedule and 6\% of the training steps for warm-up. The learning rate is set to 1e-4. The other hyper-parameters are kept the same with \cite{wolf-etal-2020-transformers}. In every 200 steps, the model is evaluated on the dev set. Patience is set to 10, that is, if the model does not achieve a lower development set loss for 10 evaluation runs, the training stops. The best checkpoint on the dev set is used to run predictions on the test set.

\section{Ablation on the ALoRA framework} 
\label{sec:appendix_ablation_alora}

We consider two variants of ALoRA: (a) use the architectural weights $\alpha_{m, i}^{'}$ as the importance scores during bi-level optimization \cite{Liu2019DARTSDA}. This variant is denoted as ALoRA-DNAS. (b) Use the sensitivity-based metric in \citet{Zhang2023AdaptiveBA} as the importance measurement. (denoted as ALoRA-Sensi). The experiments on the BoolQ and E2E methods are provided in \ref{tab:results_ablation_alora}

\begin{table}[tb!]
\centering
\resizebox{0.42\textwidth}{!}{
\begin{tabular}{c|ccc}
\hline
\multirow{2}*{\textbf{Method}}    &     \textbf{BoolQ}     &   \textbf{ReCoRD}   &    \textbf{Squad}  \\ 

&    \textbf{(acc)}  &   \textbf{(f1-em)}   &   \textbf{(f1-em)}  \\
\hline
ALoRA-DNAS   &     87.6  &  91.2  &  88.7   \\
ALoRA-Sensi   &    87.5   &  91.3  &  88.6   \\
ALoRA   &   88.0  &  91.8   &  89.2   \\

\hline
\end{tabular}}

\caption{\label{tab:results_ablation_alora} The comparison of ALoRA's variants on the BoolQ, ReCoRD, and Squad tasks. The backbone model is LlaMA-2 7B. } 
\end{table}

\section{Ablation on the pretrained backbones}

Our main experiments are conducted on the LlaMA-2 7B model. To demonstrate that our method works well regardless of the backbone models, we now conduct experiments on the RoBERTa-large. In this experiment, since the language modeling capabilities of these RoBERTa-large can not match LlaMA-2 7B model, we change the following setting for the prediction head: (a) we use a linear layer as the prediction head for classification tasks. (b) for the ReCoRD task, we use two linear layers to predict the starting and ending positions of a entity span. The other experimental settings are kept the same with the main experiments (Table \ref{tab:results_main_1}). 

We conduct experiments on the BoolQ, ReCoRD and Squad tasks. The results are reported in Table \ref{tab:results_roberta_large}. We can see that on the RoBERTa-large backbone, our method can also outperform the baseline methods.

\begin{table*}[tb!]
\centering
\resizebox{0.7\textwidth}{!}{
\begin{tabular}{c|cc|ccc}
\hline
\multirow{2}*{\textbf{Method}}   &   \multicolumn{2}{c}{\textbf{Additional Params}}    &     \textbf{BoolQ}     &   \textbf{ReCoRD}   &    \textbf{Squad}  \\ 

&  \textbf{Initial}  &  \textbf{Final}   &  \textbf{(acc)}  &   \textbf{(f1-em)}   &   \textbf{(f1-em)}  \\
\hline

\multicolumn{6}{c}{\textbf{\emph{Results for RoBERTa-large}}}  \\
\hline
Learned-Adapter   &   366M    &    354M   &  86.8   &    90.2   &  88.7   \\
\hdashline
LoRA   &    3.54M &  3.54M   & 86.9    &   90.0 
 &   88.6  \\
SoRA   &    708M  &  3.53M   &   87.2   &  90.1 
  &  88.7  \\
ALoRA   &   3.54M   &  3.42M   &  87.6   &   90.7   &    89.4  \\

\hline
\end{tabular}}

\caption{\label{tab:results_roberta_large} The comparison on the BoolQ, ReCoRD, and Squad tasks, when the backbone model is RoBERTa-large. We report the median performance over 5 random seeds. Bold and Underline indicate the best and the second best results. The metric for each task is explained in Appendix \ref{sec:appendix_evaluations}.} 
\end{table*}

We also run GPT2-large on the E2E task, and the results are reported in Table \ref{tab:results_e2e_gpt2}. The results demonstrate that when the GPT2-large is the backbone model, our ALoRA method also outperforms the baselines.

\begin{table}[tb!]
\centering
\resizebox{0.47\textwidth}{!}{
\begin{tabular}{c|ccc}
\hline
\textbf{Method}   &    \textbf{BLEU}   &   \textbf{ ROUGE-L}   &   \textbf{METEOR}    \\ 
\hline

Learned-Adapter   &  68.6    &    69.6    &   45.2   \\
LoRA   &     68.7   &  69.8   &  45.3     \\
SoRA   &   \underline{68.9}   &   \underline{69.9}   &    \underline{45.4}   \\
\hdashline
ALoRA   &   \textbf{69.3}   &  \textbf{70.4}  &     \textbf{46.0}     \\

\hline
\end{tabular}}
\caption{\label{tab:results_e2e_gpt2} Results for different PEFT methods on the E2E benchmark. The backbone LM is GPT2-large. The metrics are explained in Appendix \ref{sec:appendix_evaluations}.}
\end{table}

\section{Visualization of the final rank allocations of SoRA }
\label{sec:appendix_visualize}

\begin{figure}
\centering
\includegraphics[width=0.48\textwidth]{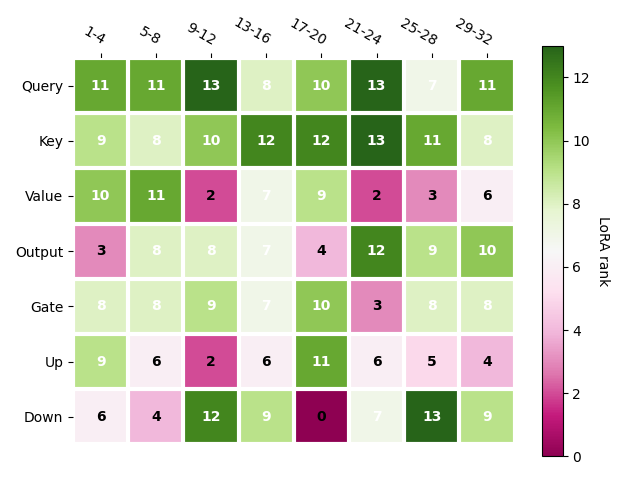}
\caption{The final rank allocations of SoRA after fine-tuning the LlaMA-2 7B model on the E2E task.}
\label{fig:visualize_sora}
\end{figure}

In the main contents, we visualize the final rank allocations of ALoRA after the training process on the E2E task in Figure \ref{fig:visualize}. As comparison, we now present the LoRA rank allocations by the SoRA method in Figure \ref{fig:visualize_sora}.

\section{Case studies of Instruction tuning}
\label{sec:appendix_case_study}

In the Section \ref{subsec:main_results} of the main content, we present the overall performance of ALoRA and SoRA on the MT-Bench, after fine-tuning LlaMA-2 7B on the Alpaca dataset. Now we present concrete examples in Table \ref{tab:results_alpaca_examples_1} to showcase the Superiority of ALoRA.

\begin{table*}[tb!]
\centering
\resizebox{0.98\textwidth}{!}{
\begin{tabular}{p{2.0cm}|p{12cm}}
\hline

Query &   \makecell[l]{Edit the following paragraph to correct any grammatical errors: \\She didn't remembre where is her purse, so I thinks its in the car but he's say \\it's on kitchen table but he are not sure, and then they asked me to looking for\\ it, she's say, "Can you?", and I responds with, "Maybe, but ain't no sure," and \\he not heard me, and, "What?", he asks, "Did you found it?".} \\
\hline

\multicolumn{2}{c}{\emph{ALoRA's response}}  \\
\hline
Responses  &   \makecell[l]{She didn't remember where her purse was, so I thought it was in the car, but he \\said it was on the kitchen table, but he wasn't sure, and then they asked me to \\look for it, she said, "Can you?" and I responded with, "Maybe, but I'm not \\sure," and he didn't hear me, and then he asked, "Did you find it?"}  \\  
\hdashline
GPT-4's comments     &  \makecell[l]{Assistant 1's response effectively corrects all the grammatical errors in the \\original paragraph. The assistant has correctly changed "remembre" to \\"remember," rearranged the clause order from "where is her purse" to "where \\her purse was," and adjusted verb forms for proper agreement (e.g., changing \\"thinks" to "thought"). Quotation marks and commas are properly used, making \\direct speech clear and punctuated correctly. The conjunctions are better placed, \\contributing to a more natural flow of sentences. Additionally, all instances of \\colloquial or incorrect language such as “ain't no sure” have been revised into \\standard English ("I'm not sure"). Overall, this response would be very helpful \\to human users as it provides a grammatically sound version of the original text.}   \\
\hdashline
GPT-4 score   &    9   \\
\hline

\multicolumn{2}{c}{\emph{SoRA's response}}  \\
\hline
Responses  &   \makecell[l]{She didn't remember where her purse was, so I thought it was in the car, but he \\said it was on the kitchen table, but he wasn't sure. They asked me to look for it,\\ and she said, "Can you?" I responded with, "Maybe, but I'm not sure." He didn't \\hear me, and he asked, "What?" I said, "Did you find it?"}  \\  
\hdashline
GPT-4's comments     &  \makecell[l]{Assistant 2 also does an excellent job correcting grammatical mistakes from the \\initial query. This assistant similarly fixes issues with verb tense consistency and \\subject-verb agreement (e.g., “he's say” corrected to “he said”). However, there \\is a slight deviation at the end where Assistant 2 writes: “I said, 'Did you find \\it?'”. This seems like an attempt by Assistant 2 to make sense of a confusing \\part of dialogue in which speaker roles were unclear but results in altering the \\meaning slightly because it suggests that ‘I’ am asking if ‘he’ found it rather \\than him asking ‘me’. This could be misleading or confusing without additional \\context indicating who is speaking at that moment. Nevertheless, punctuation \\around quotations is accurate which aids readability.}   \\
\hdashline
GPT-4 score   &    8   \\
\hline

\end{tabular}}
\caption{\label{tab:results_alpaca_examples_1} An example showcasing that the fine-tuning quality of ALoRA is higher than SoRA.  }
\end{table*}

\end{CJK*}

\end{document}